\documentclass[10pt,twocolumn,letterpaper]{article}

\usepackage[pagenumbers]{iccv} 

\definecolor{iccvblue}{rgb}{0.21,0.49,0.74}
\usepackage[pagebackref,breaklinks,colorlinks,allcolors=iccvblue]{hyperref}
\usepackage{amsmath}
\usepackage{enumitem}
\usepackage{amssymb}
\usepackage{pifont}
\usepackage{booktabs}
\usepackage{upgreek}
\usepackage{bm}
\def\paperID{9058} 
\def\confName{ICCV}
\def\confYear{2025}

\title{GAS: Generative Avatar Synthesis from a Single Image}

\author{
Yixing Lu\textsuperscript{1} \quad
Junting Dong\textsuperscript{2}\textsuperscript{\dag} \quad
Youngjoong Kwon\textsuperscript{3} \quad 
Qin Zhao\textsuperscript{2} \quad \\
Bo Dai\textsuperscript{2} \quad 
Fernando De la Torre\textsuperscript{1} \\
\textsuperscript{1}Carnegie Mellon University \quad
\textsuperscript{2}Shanghai AI Laboratory \quad
\textsuperscript{3}Stanford University \\
\url{https://humansensinglab.github.io/GAS/}
}

\begin{document}
\twocolumn[
    \maketitle
    \begin{center}
    \vspace*{-15pt}
    \includegraphics[width=\textwidth]{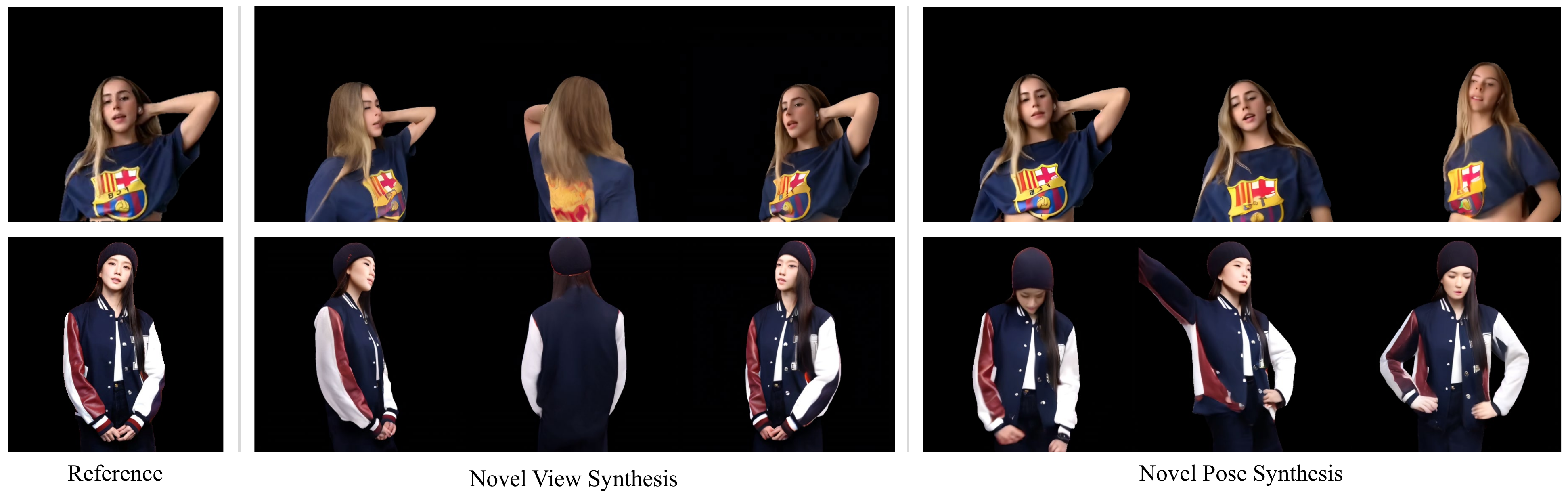}
    \vspace*{-20pt}
    \captionof{figure}{
    \textbf{In-the-wild avatar synthesis across views and poses.} Starting from a reference image, we generate its novel views and animate the avatar given a pose sequence.
    }
    \label{fig:teaser}
\end{center}
    \bigbreak
]

\begin{abstract}
We present a unified and generalizable framework for synthesizing view-consistent and temporally coherent avatars from a single image, addressing the challenging task of single-image avatar generation.
Existing diffusion-based methods often condition on sparse human templates (e.g., depth or normal maps), which leads to multi-view and temporal inconsistencies due to the mismatch between these signals and the true appearance of the subject.  Our approach bridges this gap by combining the reconstruction power of regression-based 3D human reconstruction with the generative capabilities of a diffusion model. In a first step, an initial 3D reconstructed human through a generalized NeRF provides comprehensive conditioning, ensuring high-quality synthesis faithful to the reference appearance and structure.  Subsequently, the derived geometry and appearance from the generalized NeRF serve as input to a video-based diffusion model. This strategic integration is pivotal for enforcing both multi-view and temporal consistency throughout the avatar's generation. Empirical results underscore the superior generalization ability of our proposed method, demonstrating its effectiveness across diverse in-domain and out-of-domain in-the-wild datasets.

\end{abstract}
\vspace{-1.5ex}    
\section{Introduction}
\label{sec:intro}

Human avatar generation has been a longstanding focus in computer vision and graphics, with wide-ranging applications in gaming, film, sports, fashion, and telepresence. However, despite its transformative potential, existing technologies often rely on expensive capture systems and complex workflows, making them inaccessible to the broader public. Recent advancements have aimed to make avatar generation more accessible by leveraging neural rendering techniques. In particular, studies such as~\cite{kwon2021neural, kwon2023neural, hu2023sherf, kwon2024generalizable} explore generalizable 3D human reconstruction, enabling novel view and arbitrary pose synthesis of  subjects from highly sparse—sometimes even single—input images. These approaches incorporate 3D human priors to support accurate synthesis of complex body geometry and smooth interpolation between input observations. However, their regression-based nature limits their extrapolation capabilities, often resulting in blurry outputs and restricting them to mostly rigid deformations.

\renewcommand{\thefootnote}{}
\footnotetext{\textsuperscript{\dag} Corresponding author}

\setcounter{footnote}{0} 
\renewcommand{\thefootnote}{\arabic{footnote}}
Generative models, such as GANs and diffusion models, have recently demonstrated impressive capabilities in synthesizing photorealistic images and videos with realistic motion. Leveraging these advancements, recent approaches employ diffusion models conditioned on human priors—such as 2D keypoints, depth, or normal maps—to generate high-quality avatars from a single image ~\cite{hu2023animateanyone, zhu2024champ, chang2023magicpose, shao2024human4dit}. However, the sparsity of these conditioning cues often leads to inconsistencies in the generated results, including flickering across views and temporal instability.

To address these challenges, we introduce a method for single-image avatar synthesis that ensures both view and temporal consistency. Rather than depending solely on sparse conditioning signals (e.g., SMPL normal maps), we first generate intermediate novel views and/or poses using a regression-based 3D human reconstruction model.
The shape and apperance of these new views/poses are then used as inputs to a video diffusion model. By leveraging the dense structural information from the 3D reconstruction, our approach maintains geometric fidelity and visual richness, delivering high-quality, consistent results across viewpoints and human poses (i.e., time). 

Another major challenge lies in improving generalization to real-world data, as the ultimate goal is to develop a system that performs well beyond controlled environments—specifically on casually captured, in-the-wild human imagery. Most existing human digitization models are trained on multi-view datasets collected in studio settings~\cite{kwon2021neural, kwon2023neural, hu2023sherf}, which lack the diversity found in real-world scenes, including variations in lighting, clothing, and movement. To overcome this limitation, we propose leveraging in-the-wild internet videos, which offer a rich and diverse distribution of real-world appearances. However, training with such data introduces new challenges, particularly for novel view synthesis, which traditionally relies on multi-view supervision—an element typically absent in internet-sourced footage. 

Our method, Generative Avatar Synthesis (GAS) is a unified framework that jointly learns both novel view and pose synthesis by sharing parameters across tasks. While studio-captured multi-view datasets are used for view synthesis, we augment training with both multi-view and in-the-wild videos for pose synthesis. This parameter sharing allows improvements from pose generalization to transfer naturally to view synthesis, significantly enhancing performance on real-world data (see Figure \ref{fig:teaser}). To enable this joint learning effectively, we incorporate a mode switcher that distinguishes between the two tasks. This design allows the network to prioritize view consistency for novel view synthesis and realistic deformation for novel pose generation.

In summary, our main contributions are:
\begin{itemize}
\item A unified framework for novel view and pose synthesis of avatars, which enables shared model parameters across both tasks with real human data (e.g., internet videos) at scale for training, leading to broad generalizability.
\item A dense appearance cue derived from generalizable NeRF renderings as diffusion guidance, ensuring consistent appearance preservation over novel views and poses.
\end{itemize}

\section{Related Work}
\label{sec:related_work}

\subsection{Generalizable Human Radiance Fields}

Radiance field methods like NeRF~\cite{mildenhall2021nerf} have demonstrated impressive performance in generating high-fidelity novel views. To extend these models to generalize across scenes and work with sparse-view inputs, several approaches~\cite{yu2021pixelnerf, saito2019pifu, wang2022attention} incorporate pixel-aligned features. However, applying them directly to human modeling remains challenging due to complex body geometry and self-occlusions, often resulting in over-smoothed outputs. To address this, recent works~\cite{kwon2021neural, kwon2023neural, Zhao_2022_CVPR, peng2021animatable, peng2021neural, dong2022totalselfscan, dong2023ivs} leverage 3D human priors such as the SMPL model~\cite{loper2015smpl} to better anchor features on the human form, enabling robust synthesis from sparse or even single-view inputs~\cite{hu2023sherf}. However, these methods typically suffer from slow rendering speeds.

To overcome this, 3D Gaussian Splatting~\cite{kerbl20233d}, accelerated by GPU rasterization, has emerged as an efficient radiance field representation. Recent works~\cite{zheng2024gps, kwon2024generalizable, zhuang2024idolinstant} leverage this framework for photorealistic human rendering from sparse inputs. Yet, they still face difficulties in generating fine details in unseen regions due to the mean-seeking nature of one-to-many mappings~\cite{liu2021neural, kwon2024deliffas}. Moreover, animating avatars typically requires costly post-processing—e.g., using linear blend skinning~\cite{lewis2000pose}—which often fails to capture non-rigid effects such as garment dynamics~\cite{zhuang2024idolinstant, hu2023sherf, kwon2021neural, qiu2024anigs, huang2024tech, wang2025holigs}.

In this work, we propose a novel framework that learns a distribution shift over generalizable human radiance fields, enabling consistent and artifact-free synthesis of novel views and poses. By incorporating a strong generative prior, our method naturally produces realistic non-rigid deformations from pose sequences without relying on complex post-processing pipelines.

\subsection{Generative Human Animation}
Generative human animation aims to employ generative models to produce coherent videos from static human images, utilizing guidance such as text and motion sequences. This body of work focuses on leveraging generative priors to sample complex dynamic motions, including pose-dependent clothing deformations. Early approaches harnessed the generative capabilities of Generative Adversarial Networks (GANs) \cite{goodfellow2014generative} for synthesizing novel human poses \cite{wang2021one, chan2019dance, liu2019neural}. In recent years, latent diffusion models \cite{rombach2022high} have gained traction in the realm of human animation due to their robust controllability and superior generation quality. Various methods \cite{wang2024disco, chang2023magicpose, hu2023animateanyone, zhu2024champ, li2024dispose, sun2024drive, chang2025x, li2024synthesizing} implement distinct motion guidance and conditioning techniques. Notably, Animate Anyone \cite{hu2023animateanyone} introduces a UNet-based ReferenceNet to extract features from reference images, utilizing DWPose \cite{yang2023effective} for pose guidance. Subsequent works \cite{zhu2024champ} also incorporate guidance from 3D human parametric models, such as SMPL \cite{loper2015smpl, zhao2024metric,zhao2024synergistic}, leveraging the advantages of multiple forms of guidance. Following this trajectory, recent studies \cite{he2024magicman, shao2024human4dit, qiu2024anigs, liu2024human,kant2025pippo, yang2025sigman, li2025pshuman} explore human view controllability within the diffusion frameworks. Human4DiT \cite{shao2024human4dit} develops a hierarchical 4D diffusion transformer that disentangles the learning of 2D images, viewpoints, and time. 
However, these methods struggle to synthesize view-consistent and temporally coherent results due to the gap between the sparse driving signal and the actual subject.
In this paper, we address this challenge by densely conditioning on generalizable geometry and appearance cues, leading to improved appearance preservation and consistency.

\subsection{Diffusion Models for Video Generation}
Diffusion models have shown impressive results in image synthesis and are now being adapted to the more complex domain of video generation~\cite{zeng2024dawn}. A common strategy involves extending UNet-based image diffusion models by adding temporal modules. Some approaches~\cite{blattmann2023align, guo2023animatediff} freeze the pre-trained image backbone and train only the temporal components on video data. Stable Video Diffusion (SVD)~\cite{blattmann2023stable}, for instance, enhances the UNet architecture by inserting temporal layers after each spatial convolution and attention block, and fine-tunes the full model on large-scale curated video datasets. This approach has proven to be a powerful foundation, enabling downstream applications such as video generation and 3D/4D synthesis~\cite{voleti2024sv3d, xie2024sv4d}.
In parallel, diffusion transformers~\cite{peebles2023scalable, yang2024cogvideox, videoworldsimulators2024} have emerged as an alternative architecture, applying full spatio-temporal attention over latent codes for both images and videos. In this work, we build on the strong capabilities of SVD to model multi-view consistency and temporally coherent, realistic deformations of human subjects.

\section{Method}
\label{sec:method}

\begin{figure*}[t]
    \centering
    \includegraphics[width=\textwidth]
    {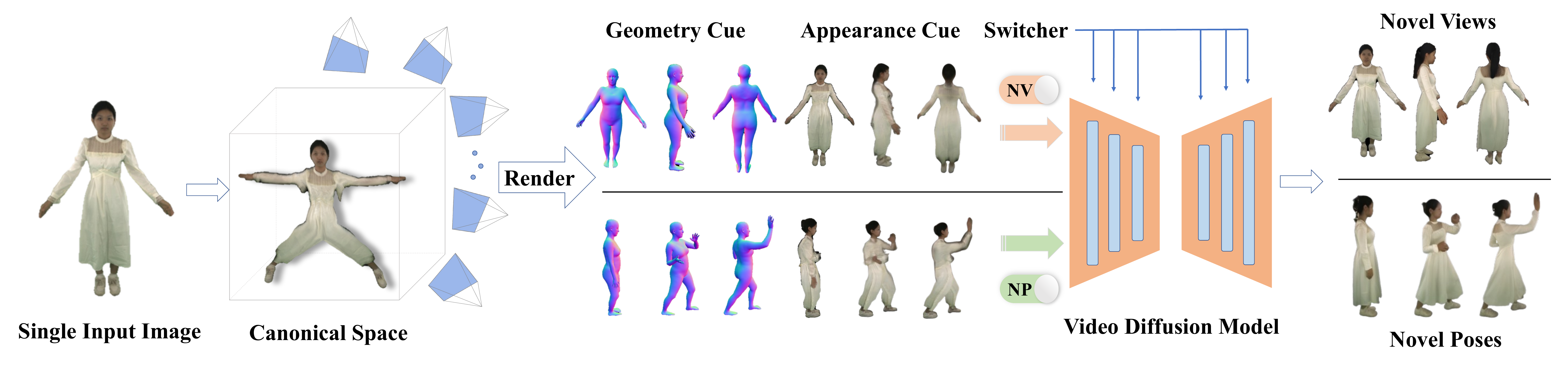}
    \caption{\textbf{Overview of GAS.} 
    Starting from a single input image, GAS uses a generalizable human NeRF to map the subject into a canonical space, then reposes and renders the 3D NeRF model to extract detailed appearance cues (i.e., NeRF renderings).
    These are paired with geometry cues (i.e., SMPL normal maps) and fed into a video diffusion model. A switcher module disentangles the tasks, enabling the model to generate either multi-view consistent novel views or temporally coherent pose animations.}
    \label{fig:pipeline}
    \vspace{-1.5em}
\end{figure*}

Figure \ref{fig:pipeline} illustrates the main idea of our method. 
 GAS synthesizes view- and pose-consistent avatar renderings from a single image by combining the rich appearance information from a generalizable 3D human reconstruction model with the generative power of a video diffusion model.

\subsection{Notation}
%
Functions (e.g., neural network mapping) are denoted with uppercase calligraphic letters (e.g., $\mathcal{U}$). Vectors are denoted with bold lowercase letters (e.g., $\bm{x}$). Matrices are denoted with uppercase letters (e.g., $C$). Sets are denoted with bold uppercase letters (e.g., $\bm{I}_{nerf}$).

\subsection{Single-view Generalizable Human Synthesis}
Given a single reference image $I_{\text{ref}}$, the camera parameter $P_\text{ref}$, and the parameters of a human template, \textit{i.e.}, the SMPL~\cite{loper2015smpl} model, we adopt single-view generalizable human NeRF \cite{hu2023sherf} to synthesize an image corresponding to the target camera parameters $P_{tar}$ and target template parameters consists of pose $\bm{\theta}_{tar}$ and shape $\bm{\beta}_{tar}$. 

To render the target image, a point $\bm{x}$ is sampled along the cast rays in the target space. Then it is transformed to point $\bm{x_c}$ in the the SMPL canonical space via inverse LBS, where the features associated with $\bm{x_c}$ are queried from the observation space. Specifically, pixel-aligned features and human template-conditioned features are obtained and fused together, denoted as $\bm{p}$. The density $\bm{\sigma}$ and color $\bm{c}$ are obtained by a multi-layer perception
(MLP) network $\mathcal{F}$:
\begin{equation}
    \bm{\sigma}(\bm{x}), \bm{c}(\bm{x}) = \mathcal{F}(\bm{x}, \bm{p}, \gamma_d(\bm{d})),
\end{equation}
where $\gamma_d$ is the positional encoding of viewing direction $\bm{d}$.

The target image  $I_{tar}$ , corresponding to the desired view and pose, is generated using volume rendering as described in~\cite{mildenhall2021nerf}. We emphasize that this human NeRF model is designed to generalize, enabling the synthesis of human appearances across a range of novel views and poses. 
%
However, due to its mean-seeking property, producing sharp renderings, particularly in occluded or unseen regions, remains challenging. 
This limitation motivates our introduction of generative priors to learn a distribution over the human views and poses, as discussed in the next section.

\subsection{Synthesis as Video Generation Condition}
This section describes how we leverage a video diffusion prior to reformulate novel view and pose synthesis as a unified video generation task, conditioned on human radiance fields and geometric templates. While previous methods typically treat view and pose synthesis as separate problems, our approach bridges them by conditioning the diffusion model on NeRF-rendered images. This dense and appearance-rich signal leads to more consistent outputs. In contrast, conditioning solely on sparse human templates, as in prior work~\cite{zhu2024champ}, often results in inconsistent appearance across views. 

\noindent \textbf{Novel View Synthesis.} Given a reference image $I_{ref}$ and a camera trajectory $\{P_1, P_2, \cdots, P_T\}$, we render the corresponding images $\bm{I}_{nerf} = \{I_1, I_2, \cdots, I_T\}$ from the NeRF, which servers as an input to our video diffusion model, Stable Video Diffusion (SVD)~\cite{blattmann2023stable}. SVD has a spatio-temporal attention module and 3D residual convolution in the diffusion UNet. For the single input image $I_\text{ref}$, we extract its feature with CLIP \cite{radford2021learning} and repeat it for $T$ times, denoted as $\bm{h}_\text{clip}$, which is then added to the video diffusion model through the cross attention. Meanwhile, we use the VAE encoder to encode our input image $I_\text{ref}$ and obtain its latent feature $C_\text{vae}$.

To introduce NeRF renderings $\bm{I}_{nerf}$ to our video diffusion model, we feed $\bm{I}_{nerf}$ to the VAE encoder, after which it is further encoded by a small convolutional neural network. We denote the output latent feature as $C_{\text{nerf}}$.

In practice, we observe that solely relying on NeRF renderings in our video diffusion model is insufficient, as these renderings may sometimes exhibit artifacts, particularly due to inaccurate SMPL fittings or occlusions. Such artifacts can corrupt the guidance and hinder the diffusion model from learning meaningful conditional distributions. To address this limitation, we further integrate a geometric cue, \textit{i.e.}, the SMPL model, to provide additional structural guidance. The SMPL model captures essential geometry information, which leads to the enhanced spatial consistency as well as the robust human shape recovery. 

In order to integrate this information with the NeRF rendered features in the 2D pixel space, similar to \cite{zhu2024champ}, we render the human template mesh into 2D normal maps. Specifically, we render the SMPL normal maps $\bm{M}=\{M_1, M_2, \cdots, M_T\}$ under the camera trajectory $\{P_1, P_2, \cdots, P_T\}$. Then a set of 2D convolution layers are utilized to extract the features, denoted by $C_{\text{smpl}}$.

To effectively fuse \( C_{\text{nerf}} \) and \( C_{\text{smpl}} \), we combine them through element-wise addition. The resulting fused feature is then added to the output of the first convolutional layer of the UNet in the video diffusion model.

\noindent \textbf{Novel Pose Synthesis.} Given a sequence of SMPL poses $\{\bm{\theta}_1, \bm{\theta}_2, \cdots, \bm{\theta}_T\}$ and a fixed camera parameter, we render corresponding images from the NeRF and SMPL normal maps. They can be used as conditions to the video diffusion model in the same manner as illustrated above.

Now we can formulate the learning objective of our diffusion model. The diffusion UNet \( \mathcal{U}_\Theta \) predicts the noise \( \epsilon \) for each diffusion step \( t \), and our training objective is
\begin{equation}
\mathcal{L}_{\mathcal{U}_{\theta}} = \mathbb{E} \left[ \| \epsilon - \mathcal{U}_{\theta} (Z_t, t, \bm{h}_{\text{clip}}, C_{\text{vae}}, C_{\text{nerf}}, C_{\text{smpl}}) \| \right] 
\label{eq:prev_objective}
\end{equation}

where \( Z_t = \alpha_t Z + \sigma_t \epsilon \). Here, \( Z \) is the ground-truth latent, \( \epsilon \sim \mathcal{N}(0, I) \), and \( \alpha_t \) and \( \sigma_t \) define the noise at timestep \( t \). $\Theta$ is the  learnable parameters of the UNet $\mathcal{U}$.

\subsection{Switcher for Disentangled Synthesis}
Joint learning of human view synthesis and pose synthesis presents inherent challenges for feed-forward methods. Our proposed framework addresses these challenges by unifying both tasks into a single video generation task, leveraging the capability of video diffusion model to effectively model each task under the same representation. Under this framework, a straightforward approach would be to train the video diffusion model on both multi-view and dynamic video data simultaneously. However, our empirical findings reveal that dynamic motions embedded within view synthesis videos can disrupt view consistency (see Figure \ref{fig:ablation_swicher}). This issue arises due to the inherent modality differences between static view synthesis videos and dynamic animation videos. To mitigate this problem, we introduce a switcher mechanism within our video diffusion model that explicitly controls and disentangles these two modes, ensuring task-specific consistency and performance.

In particular, we introduce the switcher \( \bm{s} \), which is a one-hot vector that labels each of the two modalities, as the additional condition to the video diffusion model. This allows us to extend the formulation in Equation \ref{eq:prev_objective} as follows:
\begin{equation}
    \mathcal{L}_{\mathcal{U}_{\theta}} = \mathbb{E} \left[ \| \epsilon - \mathcal{U}_{\theta} (Z_t, t, \bm{h}_{\text{clip}}, {C}_{\text{vae}}, {C}_{\text{nerf}}, {C}_{\text{smpl}}, \bm{s}) \| \right] 
    \label{eq:train_objective}
\end{equation}

To incorporate the domain switcher \( \bm{s} \), we first apply positional encoding and then concatenate it with the time embedding. This combined encoding is subsequently fed into the UNet within the video diffusion model.

\subsection{Training and Inference}
\noindent \textbf{Training.} The entire training process of our pipeline consists of two stages. During the first stage, we train the generalizable human NeRF $\mathcal{F}$ model on the multi-view datasets. Following \cite{hu2023sherf}, we randomly sample observation and target image pairs from each subject. The prediction of the target view is supervised by minimizing the loss objective $\mathcal{L} = \mathcal{L}_2 + \lambda_{\text{ssim}} \cdot \mathcal{L}_{\text{ssim}} + \lambda_{\text{lpips}} \cdot \mathcal{L}_{\text{lpips}} + \lambda_{\text{mask}} \cdot \mathcal{L}_{\text{mask}}
$
where $\mathcal{L}_2, \mathcal{L}_{\text{ssim}}, \mathcal{L}_{\text{lpips}}$ are photometric \(L_2\) loss, SSIM loss \cite{wang2004image} and LPIPS \cite{zhang2018unreasonable} loss between the predicted image and the ground truth. $\mathcal{L}_{\text{mask}}$ is the \(L_2\) difference between the accumulated volume density and the ground truth human binary mask. $\lambda_{\text{ssim}}, \lambda_{\text{lpips}}, \lambda_{\text{mask}}$ are weights of each loss to balance their contributions to the final loss function. 

In the second stage, we freeze the generalizable NeRF model and train our video diffusion model. We train the full spatio-temporal UNet and feature encoders for NeRF and SMPL normal renderings following our training objective in Equation \ref{eq:train_objective}. The second stage is trained on our complete dataset to ensure the generalization.

\noindent \textbf{Inference.} We apply classifier-free guidance (CFG) \cite{ho2022classifier} to inference from the video diffusion model. The two tasks are done with different CFG schedules according to the task properties. For the novel view synthesis task, we utilize a triangular CFG scaling \cite{voleti2024sv3d}, where we linearly increase CFG from 1 at the front view to 2 at the back view, then linearly decrease it back to 1 at the front view. For the novel pose synthesis task, we fix the CFG scale to be 2.

\section{Experiments}
\label{sec:experiment}
\subsection{Experimental Setup}
\subsubsection{Datasets}

\noindent\textbf{3D Scans.} We use THuman2.1 \cite{tao2021function4d} and 2K2K dataset \cite{han2023Recon2K} with around 4500 3D scans in total. THuman2.1 comprises 2445 high-quality 3D scans and texture maps. We randomly sample 2345 subjects for training and the remaining 100 subjects for testing. 2K2K dataset consists of 2000 train and 50 test subjects, totally 2050 scans. For both datasets, RGB images are rendered from 20 uniformly distributed views around the scan, at the resolution of $1024\times1024$. SMPL parameters are estimated using off-the-shelf method for multi-view SMPL fitting \cite{easymocap}.

\noindent\textbf{Multi-view Videos.} MVHumanNet \cite{xiong2024mvhumannet} is a multi-view video dataset featuring a large number of diverse identities and everyday clothing. We use a subset of 944 human captures, each consisting of synchronized 16-view videos per subject. We reserve 48 subjects for evaluation and use the remaining subjects for training. The dataset also includes SMPL parameters, optimized from multi-view images.

\noindent\textbf{Monocular Videos.} For in-the-wild datasets, we use the TikTok dataset \cite{jafarian2021learning} and an additional collection of internet videos. The TikTok dataset includes 350 dance videos, from which we processed and filtered 289 valid video sequences for training. Following the protocol in \cite{wang2024disco, chang2023magicpose}, we use subjects from 335 to 340 for testing. To further diversify the data, we selected 122 video sequences from Champ \cite{zhu2024champ} training dataset, originally sourced from reputable online platforms such as TikTok and YouTube. For both datasets, we obtain the foreground human masks using Grounded-SAM \cite{ren2024grounded} and the SMPL parameters using 4DHumans \cite{goel2023humans}. 

\subsubsection{Implementation Details}
We trained the generalizable human NeRF model \cite{hu2023sherf} on MVHumanNet dataset. 
To accelerate the video diffusion training, instead of creating and rendering the human NeRF on-the-fly, we choose to store the NeRF renderings for all datasets offline. 
For the video diffusion model, we initialize it with the pre-trained Stable Video Diffusion 1.1 image-to-video model \footnote{https://huggingface.co/stabilityai/stable-video-diffusion-img2vid-xt-1-1} \cite{blattmann2023stable}. We resize all images to a resolution of 512$\times$512. Each batch consists of 20 frames. We train the model for 150k iterations with an effective batch size of 8 and a learning rate of $10^{-5}$. We utilize 8 A100 GPUs and the total training time is 3 days. 

\subsubsection{Baselines and Metrics}
\noindent\textbf{Baselines.} We benchmark our method against state-of-the-art generative human rendering methods including Champ \cite{zhu2024champ} and Animate Anyone \cite{hu2023animateanyone}. As for Animate Anyone \cite{hu2023animateanyone}, we use the implementations from Moore Threads \footnote{https://github.com/MooreThreads/Moore-AnimateAnyone}. 

\noindent\textbf{Metrics.} 
We evaluate the fidelity and consistency of our results using both image-level and video-level metrics. For image-level comparisons, we report peak signal-to-noise ratio (PSNR), structural similarity index measure (SSIM) \cite{wang2004image}, and learned perceptual image patch similarity (LPIPS) \cite{zhang2018unreasonable}. For video-level evaluation, we use the Fréchet Video Distance (FVD) \cite{unterthiner2018towards} metric.

\begin{table*}[ht]
\centering
\resizebox{0.8\textwidth}{!}{ 
\begin{tabular}{lcccccccc}
\toprule
Method & \multicolumn{2}{c}{PSNR $\uparrow$} & \multicolumn{2}{c}{SSIM $\uparrow$} & \multicolumn{2}{c}{LPIPS $\downarrow$} & \multicolumn{2}{c}{FVD $\downarrow$} \\
\cmidrule(lr){2-3} \cmidrule(lr){4-5} \cmidrule(lr){6-7} \cmidrule(lr){8-9}
& THuman & 2K2K & THuman & 2K2K & THuman & 2K2K & THuman & 2K2K \\
\midrule
Animate Anyone & 22.48 & 18.48 & 0.927 & 0.557 & 0.061 & 0.263 & 460.3 & 1422.1 \\
Champ & 20.96 & 22.14 & 0.909 & 0.910 & 0.074 &  0.075 & 470.3 & 480.3 \\
\midrule
Animate Anyone* & 25.20 & 26.22 & 0.938 & 0.936 & 0.046 & 0.050 & 302.7 & 286.4 \\
Champ* & 23.89 & 25.66 & 0.928 & 0.935 & 0.054 &  0.052 & 296.1 & 279.3 \\
\midrule
Ours & \textbf{26.77} & \textbf{28.82}  & \textbf{0.943}  & \textbf{0.954} & \textbf{0.041} & \textbf{0.039}  & \textbf{194.8} & \textbf{191.3} \\
\bottomrule
\end{tabular}
}
\caption{\textbf{Quantitative comparison for novel view synthesis on THuman and 2K2K dataset.} For all the methods, we report the average score on 20 views using four orthogonal input views (front, back, and side views). * indicates methods fine-tuned on our 3D scan dataset.}
\label{tab:mv_comparison}
\end{table*}

\begin{figure*}[t]
    \centering
    \includegraphics[width=0.95\textwidth]{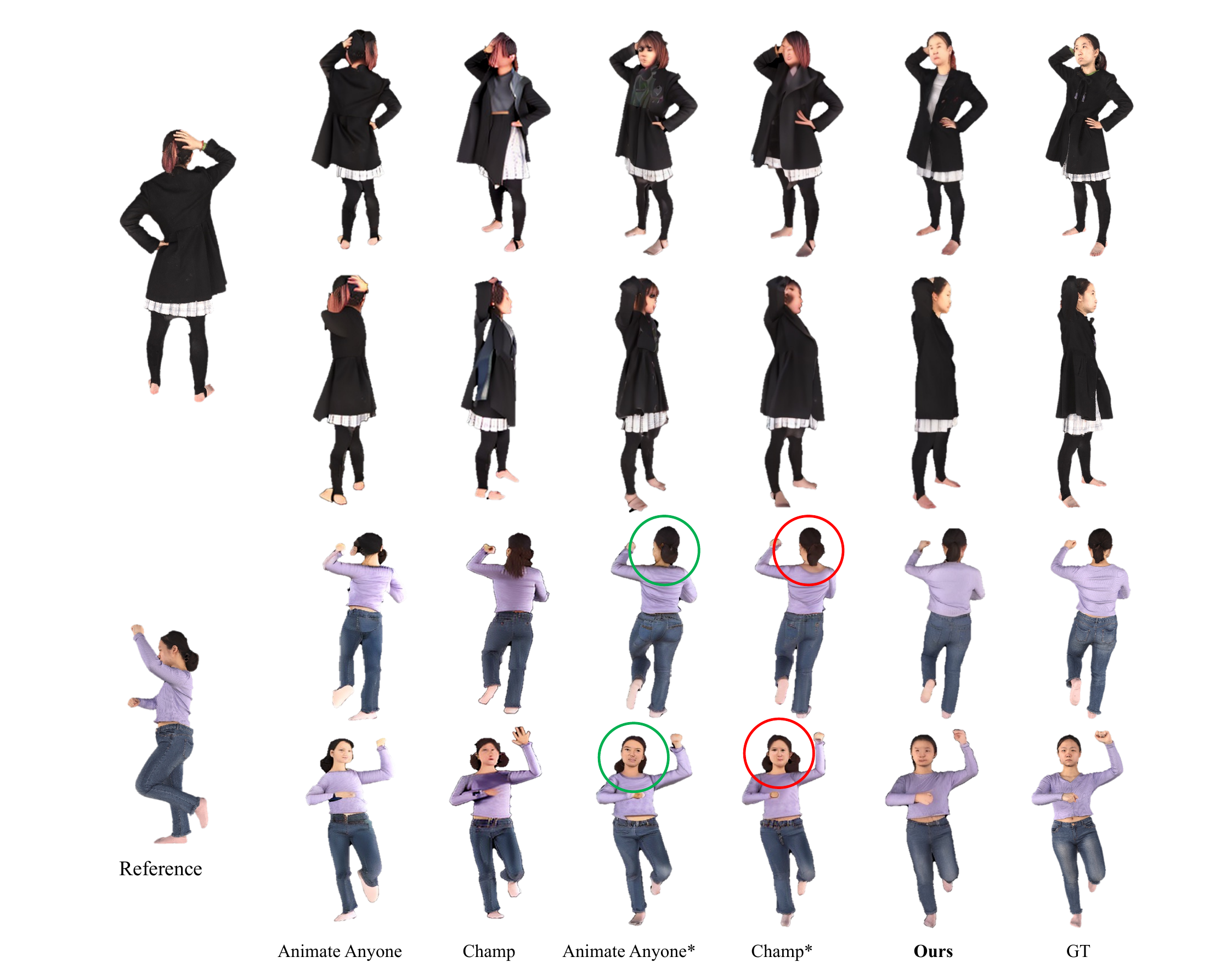}
    \caption{\textbf{Qualitative comparisons for novel view synthesis on the THuman dataset.} For the first subject, our method generates cleaner garment textures and sharper facial details, achieving better realism and consistency across views (e.g., the hair style in our generated front view is faithful to the reference image). For the second subject, baseline methods exhibit inconsistencies across views, marked with circles (e.g., hair misalignment between generated front and back views). * denotes fine-tuning on our full 3D scan dataset.}
    \label{fig:nvs_comparison}
    \vspace{-1.5em} 
\end{figure*}

\begin{figure*}[t]
    \centering
    \includegraphics[width=0.9\textwidth]{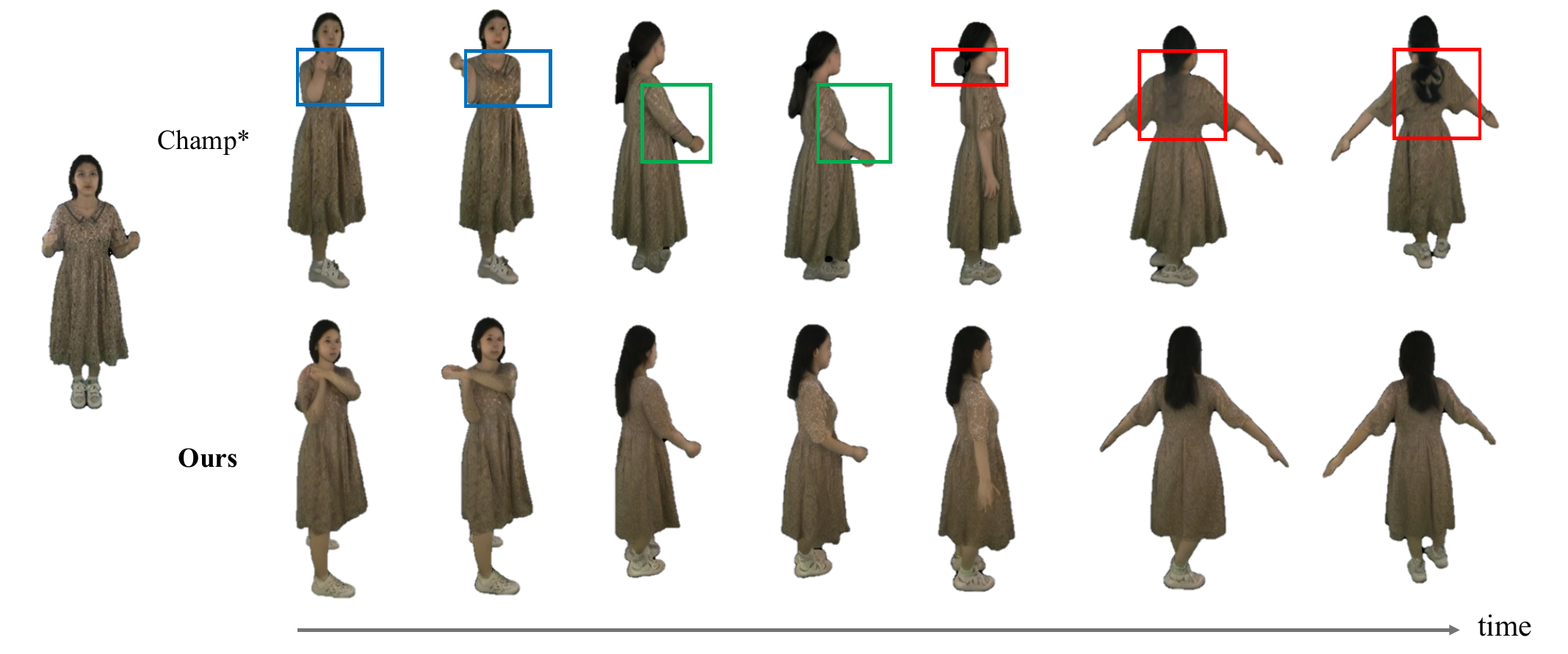}
    \caption{\textbf{Qualitative comparisons for novel pose synthesis on MVHumanNet dataset.} In the first row of Champ \cite{zhu2024champ} results, blue rectangles mark disappearing arms, green rectangles show varying sleeve lengths, and red rectangles indicate inconsistencies in hair appearance. * denotes a method further fine-tuned using our complete animation video dataset.}
    \label{fig:animation_comparison}
\end{figure*}

\subsection{Comparisons on Novel View Synthesis}
We compare our approach with two leading generative human synthesis methods—Animate Anyone~\cite{hu2023animateanyone} and Champ~\cite{zhu2024champ}. For a fair evaluation, we fine-tuned both models on our complete 3D scan dataset for 10,000 iterations.

\noindent\textbf{Quantitative results.} We employ a rigorous evaluation protocol for single-image human novel view synthesis. For each subject, we sample four orthogonal input views and report the average performance across all corresponding novel views. Notably, Animate Anyone~\cite{hu2023animateanyone} tends to generate noisy backgrounds. To ensure a fair comparison that emphasizes the synthesized human subject, we apply the ground truth masks to remove backgrounds in the THuman dataset. In contrast, for the 2K2K dataset, we evaluate the generated outputs without background removal.

Quantitative results are presented in Table \ref{tab:mv_comparison}. Results show that our method achieves state-of-the-art performance across all evaluation metrics, highlighting the advantages of our generalizable human radiance field in preserving intricate details across viewpoint changes.

\noindent\textbf{Qualitative results.} Figure \ref{fig:nvs_comparison} presents our qualitative comparisons with baseline methods, focusing on challenging reference views and novel views with less overlap.

\subsection{Comparisons on Novel Pose Synthesis}
We compare GAS on novel pose synthesis task with Animate Anyone \cite{hu2023animateanyone} and Champ \cite{zhu2024champ}. For a fair comparison, we also further fine-tuned the baseline methods on our complete video dataset for 10k iterations.

\noindent\textbf{Quantitative results.}
To evaluate on the TikTok dataset, we randomly sample a frame as the reference and generate the subsequent 100 frames.  Quantitative results are presented in Table \ref{tab:novepose_comparison}. Our method consistently outperforms the baseline methods, even after additional fine-tuning.

\noindent\textbf{Qualitative results.}
Qualitative results are shown in Figure \ref{fig:animation_comparison}, where we compare our method with Champ \cite{zhu2024champ} on the novel view animation task for the MVHumanNet dataset. Our method synthesizes temporally consistent animations, even from novel views. It highlights the advantage of leveraging human NeRF’s dense and 3D consistent appearance cue for guidance.
\begin{table}[t]
\centering
\resizebox{0.9\columnwidth}{!}{%
\begin{tabular}{lcccc}
\hline
Method & PSNR $\uparrow$ & SSIM $\uparrow$ & LPIPS $\downarrow$ & FVD $\downarrow$ \\
\hline
Animate Anyone & 17.21  & 0.762  & 0.225 & 1274.1\\
Champ & 18.48  & 0.806  & 0.182 & 585.0 \\
\midrule
Animate Anyone* & 17.83  & 0.791  & 0.204 & 840.5  \\
Champ* & 18.57  & 0.797  & 0.187 & 893.7  \\
\midrule
Ours & \textbf{19.11} & \textbf{0.833} & \textbf{0.176} & \textbf{362.0} \\
\hline
\end{tabular}
}
\caption{\textbf{Quantitative comparisons for novel pose synthesis on TikTok dataset.} * Indicates methods fine-tuned on our multi-view video and monocular video dataset.}
\vspace{-1.0em} 
\label{tab:novepose_comparison}
\end{table}

\subsection{Ablation Studies and Analyses}
We perform ablation studies to assess different variants of our proposed method. Additional quantitative and qualitative results can be found in the supplementary material.

\noindent\textbf{Generalizable geometry and appearance cues.}
To study the effect of both geometry (\textit{i.e.}, SMPL normal map) and appearance cues (\textit{i.e.}, human radiance field), we train a variant with the geometry cue completely removed and a variant with appearance cue removed. We present quantitative comparisons for novel view synthesis on the THuman dataset and novel pose animation on the MVHumanNet dataset.
As shown in Table \ref{tab:ablation_appear_cue}, our approach consistently improved performance across both tasks and datasets. 
\begin{table}[h]
\centering
\resizebox{0.9\columnwidth}{!}{%
\begin{tabular}{lcccccccc}
\toprule
Method & \multicolumn{2}{c}{PSNR $\uparrow$} & \multicolumn{2}{c}{SSIM $\uparrow$} & \multicolumn{2}{c}{LPIPS $\downarrow$} & \multicolumn{2}{c}{FVD $\downarrow$} \\
\cmidrule(lr){2-3} \cmidrule(lr){4-5} \cmidrule(lr){6-7} \cmidrule(lr){8-9}
& NVS & NPS & NVS & NPS & NVS & NPS & NVS & NPS \\
\midrule
w.o. geo. cue & 26.07 & 28.33  & 0.938  & 0.943  & \textbf{0.045}  & 0.044 & 227.1  & 210.6 \\
w.o. appear. cue & 26.38  & 27.66  & 0.938  & 0.941  & 0.041  & 0.043  & 207.5  & 234.6\\
\midrule
Ours & \textbf{26.77} & \textbf{28.74}   & \textbf{0.943}  & \textbf{0.945} & 0.041 & \textbf{0.040}   & \textbf{194.8} & \textbf{188.5} \\
\bottomrule
\end{tabular}
}
\vspace{-0.5em} 
\caption{\textbf{Quantitative ablation study on the geometry and appearance cues.} NVS denotes novel view synthesis and NPS denotes novel pose synthesis. The same notation applies to Table \ref{tab:svd}.}
\label{tab:ablation_appear_cue}
\end{table}

\begin{figure}[h]
    \centering
    \includegraphics[width=0.48\textwidth]{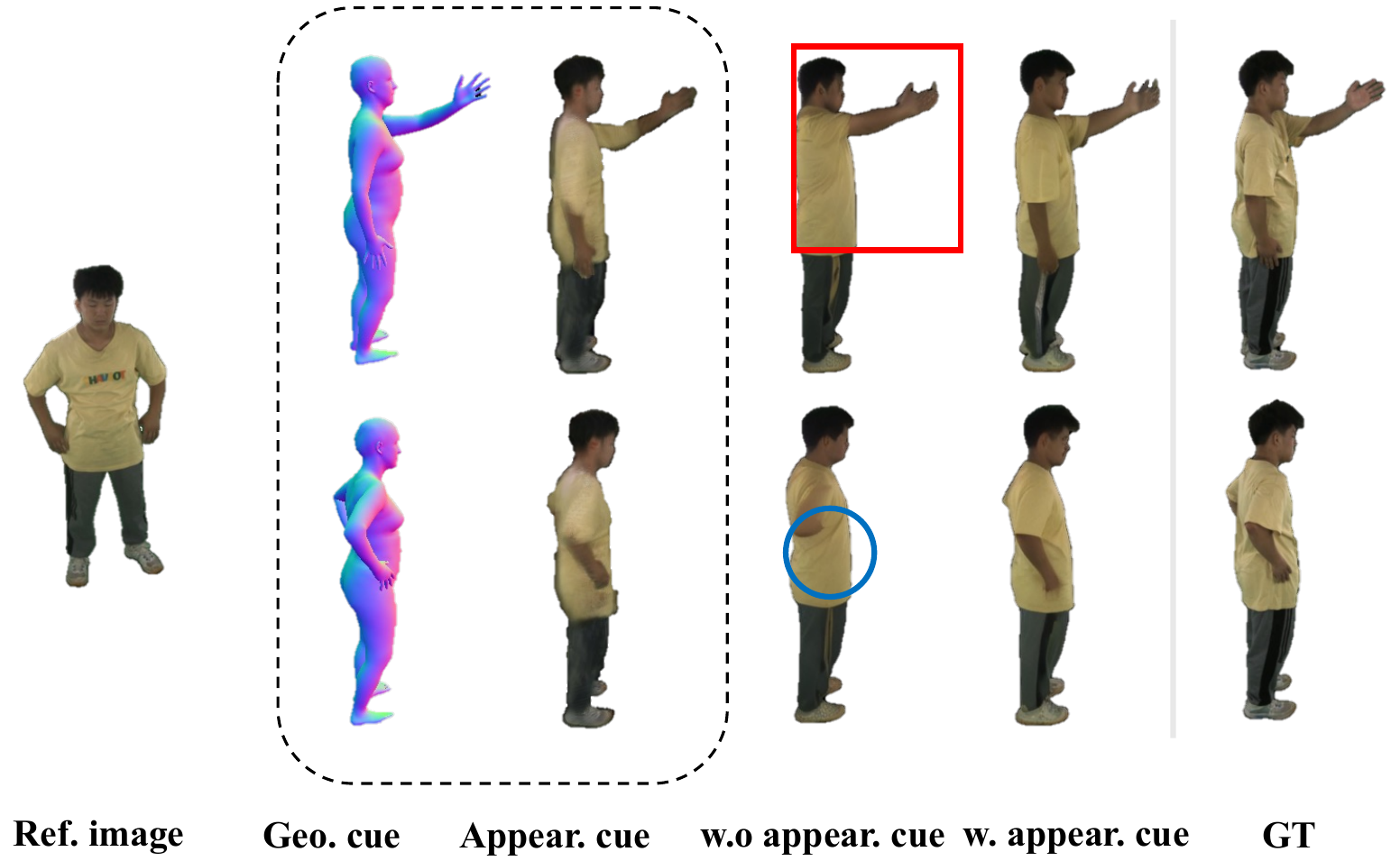}
    \caption{\textbf{Ablation study on the appearance cue.} Without the appearance cue, artifacts include incorrect arm raises (red rectangles) and distorted hand placement on the waist (blue circles), both resolved with the appearance cue.}
    \label{fig:ablation_wo_appear_cue}
\end{figure}

Our hypothesis for the generalizable human NeRF is that it provides rich and consistent appearance cues across different views and times. To validate this, we present an additional qualitative study, as suggested in Figure \ref{fig:ablation_wo_appear_cue}.

\begin{figure}[h]
    \centering
    \includegraphics[width=0.48\textwidth]{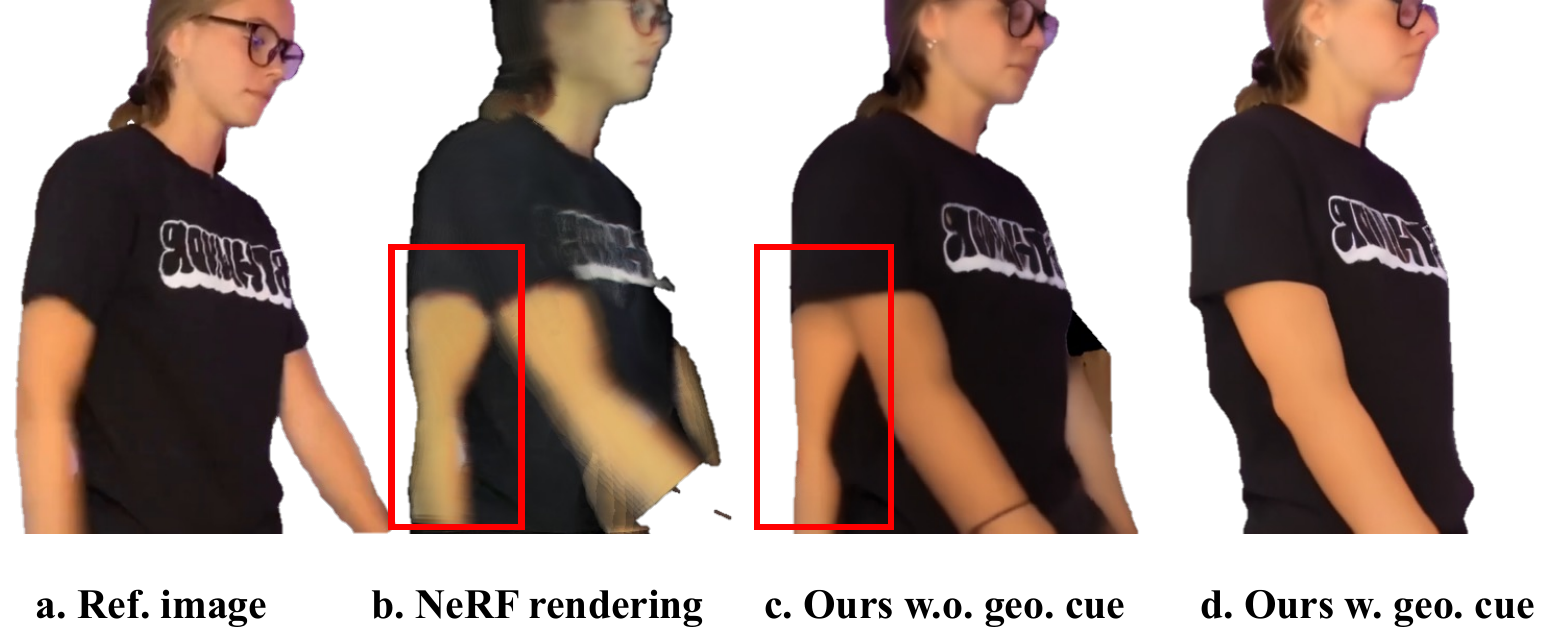}
    \caption{\textbf{Ablation study on the geometry cue.} Without the geometry cue, occlusion leads human NeRF to misinterpret the arm as clothing texture, which further misleads diffusion generation (red rectangles).}
    \label{fig:ablation_wo_geo_cue}
\end{figure}

\begin{figure}[h]
    \centering
    \includegraphics[width=0.5\textwidth]{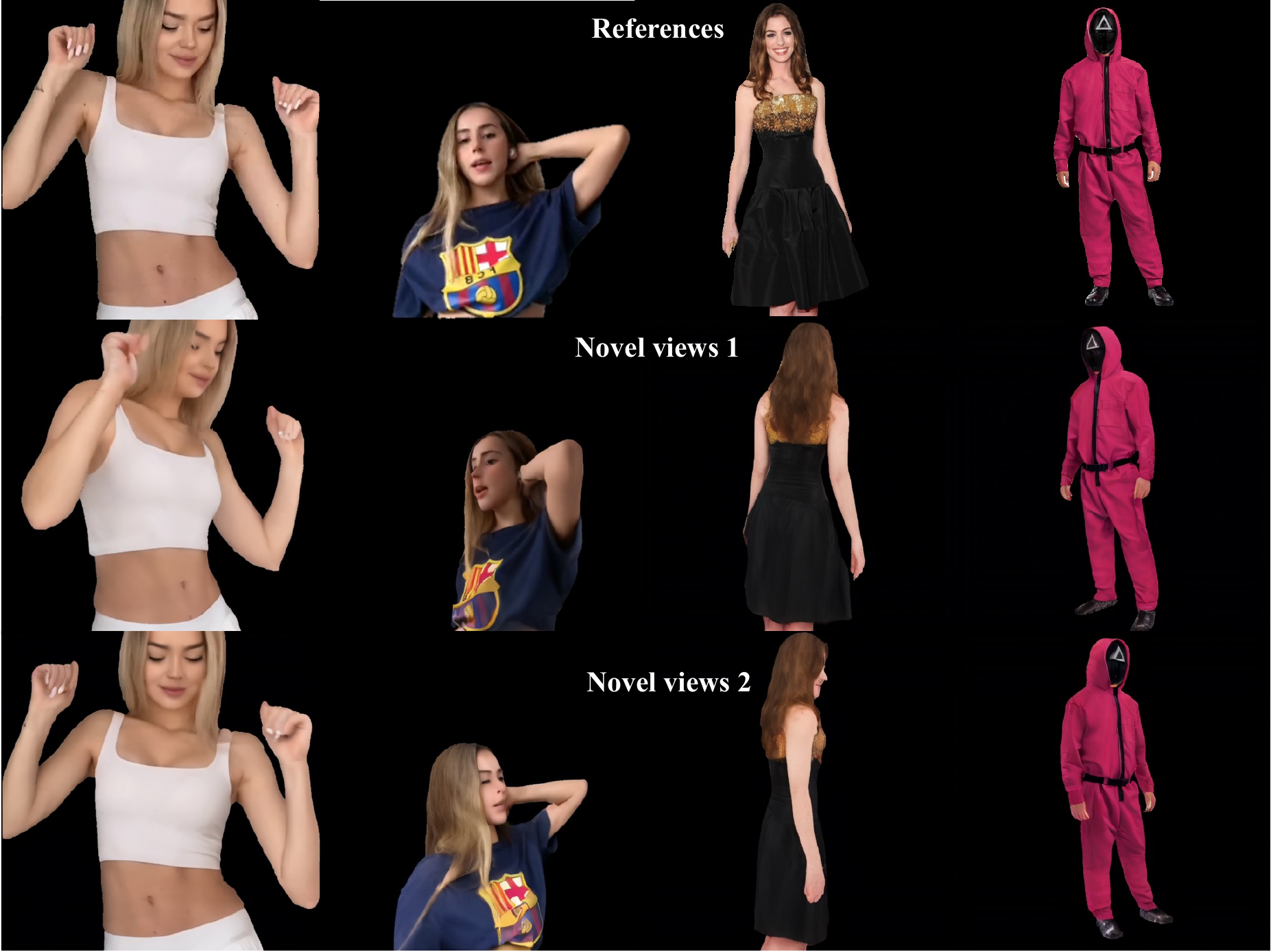}
    \caption{\textbf{Novel view synthesis on real-world human subjects.}}
    \label{fig:ablation_mixed_training}
\end{figure}

Occlusions, which are frequent in in-the-wild videos, often introduce artifacts in generalizable human NeRF due to pixel-aligned feature extraction and warping. As illustrated in Figure~\ref{fig:ablation_wo_geo_cue}, the absence of clean geometric guidance can mislead the diffusion model, resulting in noticeable distortions.

\noindent\textbf{Video diffusion model.} The video diffusion model is suggested to refine the generalizable human NeRF renderings and produce shaper results with fewer artifacts. We ablate the effect of video diffusion model in Table \ref{tab:svd} by comparing the generalizable human NeRF renderings and the diffusion refined results. Results show significant improvement across all metrics on both tasks.
\begin{table}[h]
\centering
\resizebox{0.9\columnwidth}{!}{%
\begin{tabular}{lcccccccc}
\toprule
 & \multicolumn{2}{c}{PSNR $\uparrow$} & \multicolumn{2}{c}{SSIM $\uparrow$} & \multicolumn{2}{c}{LPIPS $\downarrow$} & \multicolumn{2}{c}{FVD $\downarrow$} \\
\cmidrule(lr){2-3} \cmidrule(lr){4-5} \cmidrule(lr){6-7} \cmidrule(lr){8-9}
& NVS & NPS & NVS & NPS & NVS & NPS & NVS & NPS \\
\midrule
Before diffusion & 24.25 & 18.37   & 0.925  & 0.809   & 0.073  & 0.233  & 517.7  & 1255.8 \\
After diffusion & \textbf{28.82} & \textbf{19.11}  & \textbf{0.954}  & \textbf{0.833}  & \textbf{0.039}  & \textbf{0.176}   &  \textbf{191.3}   & \textbf{362.0} \\
\bottomrule
\end{tabular}
}
\vspace{-0.5em} 
\caption{\textbf{Quantitative ablation on the video diffusion model.}}
\label{tab:svd}
\end{table}

\begin{figure}[h]
    \centering
    \includegraphics[width=0.45\textwidth]{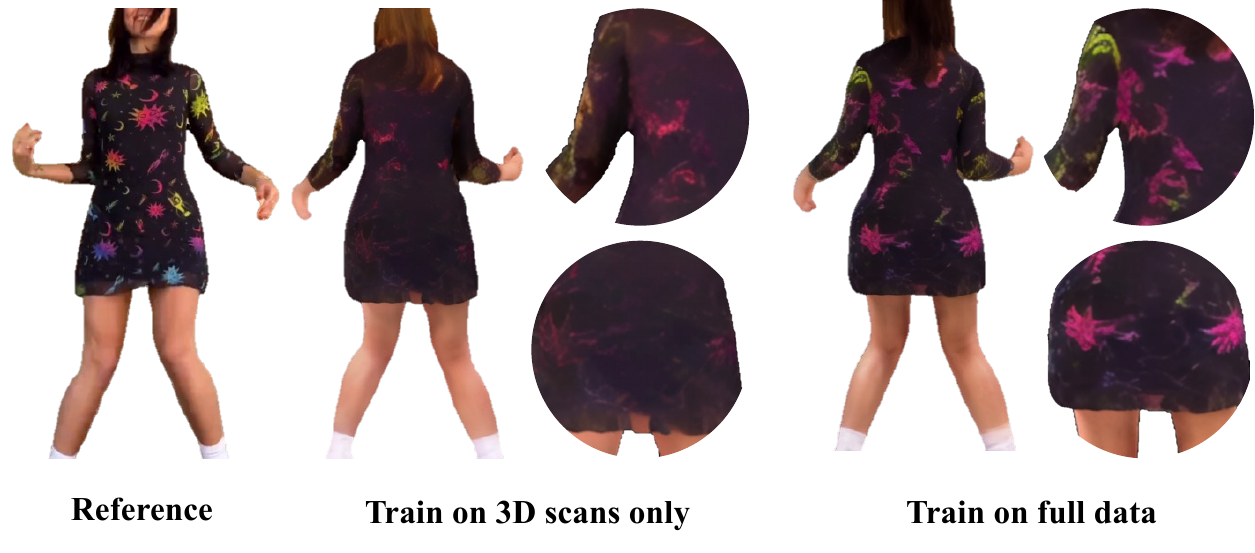}
    \caption{\textbf{Ablation on involving internet videos for training.} }
    \label{fig:ablation_generalization}
    \vspace{-1.0em} 
\end{figure}

\noindent\textbf{Mixed training with 3D/multi-view captures and internet videos.} 
We evaluate the effectiveness of our training strategy that incorporates diverse data sources to enhance generalization. As illustrated in Figure \ref{fig:ablation_mixed_training}, our model exhibits robust performance in novel view synthesis across real-world scenarios. Additionally, Figure \ref{fig:ablation_generalization} shows an ablation study highlighting the contribution of internet video data to improved generalization.

\begin{figure}[h]
    \centering
    \includegraphics[width=0.48\textwidth]{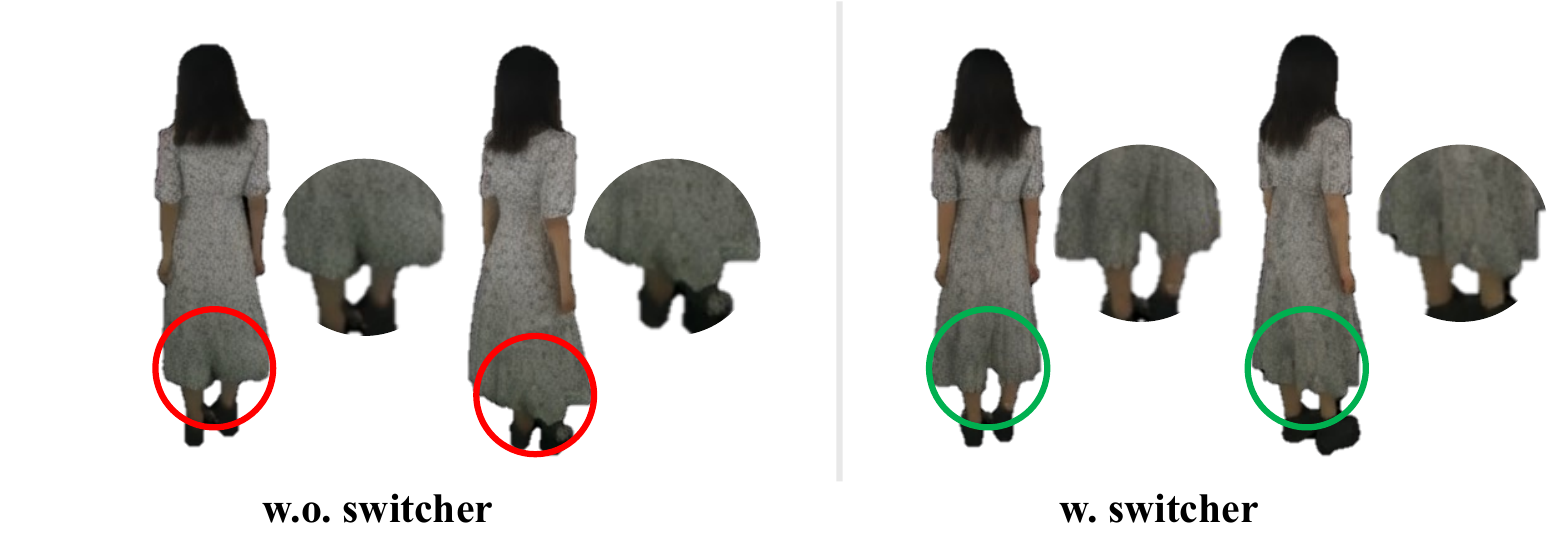}
    \caption{\textbf{Ablation on the switcher for disentangling static view and dynamic motion synthesis.} Without the switcher, undesired clothing deformations, such as dress swinging, are involved in the novel view generation, corrupting the view consistency.}
    \label{fig:ablation_swicher}
\end{figure}

\noindent\textbf{Effect of switcher for disentangled synthesis.}
In our unified framework for static view synthesis and dynamic motion modeling, we incorporate a switcher module to explicitly separate the two tasks. To assess its impact, we train a variant without the switcher. As shown in Figure~\ref{fig:ablation_swicher}, omitting the switcher leads to unintended motion artifacts in sequences of novel views, highlighting its importance for task disentanglement.

\section{Conclusion}
\label{sec:conclusion}
We propose a unified framework for avatar synthesis from a single image that tackles two major challenges: maintaining appearance consistency and generalizing to in-the-wild scenarios. Our method combines a regression-based human radiance field with a video diffusion model, leveraging dense conditioning to reduce the mismatch between driving signals and target outputs. The framework supports large-scale training on diverse, real-world human data, while cleanly separating the modeling of static novel views and dynamic motion. This combination enables high-fidelity avatar generation with realistic deformations and strong generalization across varied environments.

\section*{Acknowledgements}
This work was supported by the Shanghai Artificial Intelligence Laboratory. The authors would like to thank Aviral Chharia, Jialu Gao, Jianjin Xu, and Wenbo Gou for their valuable feedback on the manuscript.

{
    \small
    \bibliographystyle{ieeenat_fullname}
    \bibliography{main}
}

\clearpage
\newpage
\maketitlesupplementary
\section*{A Limitations and future works}
Although our method achieves state-of-the-art performance in both novel view and pose synthesis, it is not free from limitations. (1) Due to the constraints of current off-the-shelf single-image human mesh recovery methods, we use SMPL to establish geometry and appearance conditions for scalable training. However, SMPL lacks expressiveness in regions such as the face and hands, resulting in artifacts in these areas. Future works could explore scalable solutions inspired by recent efforts, such as regional supervision \cite{xu2024high}. (2) While we utilize Stable Video Diffusion to model view and pose synthesis, achieving strong consistency across both spatial and temporal dimensions, some challenges remain. Specifically, we occasionally observe degradation in fine-grained detail quality and difficulties in accurately generating complex clothing textures, particularly during significant clothing deformations. Addressing the former may involve increasing the image resolution and exploring advanced video generative models, such as CogVideoX \cite{yang2024cogvideox}. For the latter, leveraging synthetic human datasets with intricate textures presents an exciting avenue for future research.

\section*{B Ethical considerations}
\noindent\textbf{Human subjects data.} We adhere to strict ethical guidelines in the collection and use of data involving human subjects. Below, we provide details on how we obtained each dataset utilized in our work:

\begin{itemize}
    \item THuman2.1 \cite{tao2021function4d} is a publicly available dataset. We signed the necessary agreement with the dataset authors to obtain access via an official download link.
    \item 2K2K \cite{han2023Recon2K} and MVHumanNet \cite{xiong2024mvhumannet} follow the same procedure as THuman2.1, involving an agreement with the authors to obtain official download links.
    \item TikTok Dataset \cite{jafarian2021learning} is publicly available and was downloaded in its preprocessed form from MagicPose \cite{chang2023magicpose}.
    \item Additional Real-World Data was manually selected and filtered from the publicly released portions of Champ \cite{zhu2024champ} training data.
\end{itemize}

\noindent\textbf{Broader Impact.} Our proposed method enables affordable and accessible solutions for a wide range of applications. By leveraging generative AI, we transform content generation, making it possible to generate novel views and poses from just a casually-captured single image. It has the potential to revolutionize fields such as virtual reality, gaming, and digital content creation by significantly lowering the barriers to high-quality multi-view synthesis and animation.

However, along with these benefits come potential risks that warrant careful consideration. The misuse of generative AI in creating synthetic content raises ethical concerns, such as the possibility of producing misleading or harmful media. Additionally, privacy concerns may arise when real-world data is used for training, especially when the data involves human subjects. Ensuring robust safeguards, transparent practices, and compliance with ethical standards is crucial to mitigate these risks while maximizing the positive impact of our method.

\section*{C Implementation details}
\subsection*{C.1 Model architecture}

\begin{figure*}[ht]
    \centering
\includegraphics[width=\textwidth]{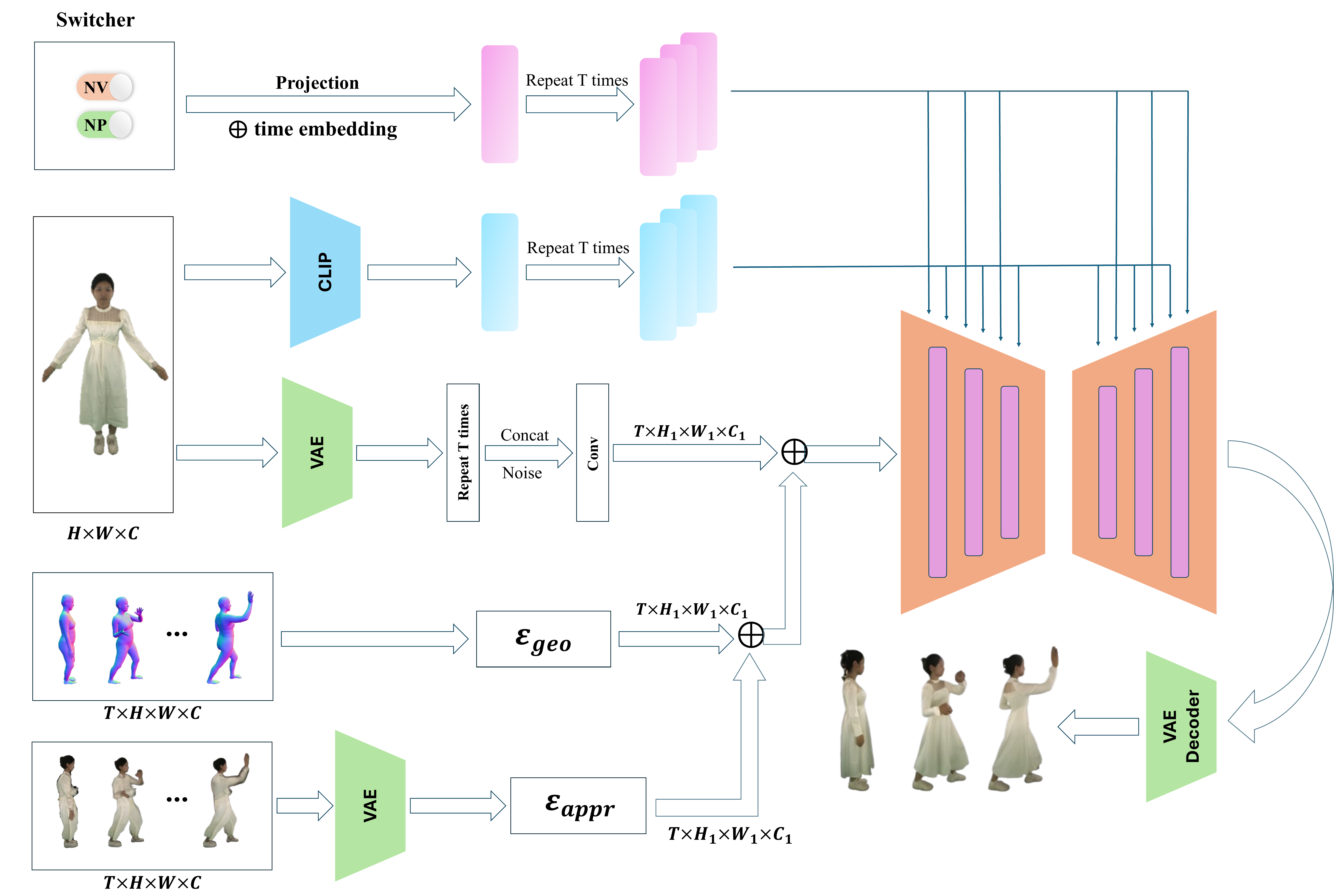}
    \caption{\textbf{Network architecture for processing the conditions of the video diffusion model.} $\boldsymbol{\epsilon}_{\text{geo}}$ and $\boldsymbol{\epsilon}_{\text{appr}}$ denote geometry encoder and appearance encoder, respectively. $\boldsymbol{\oplus}$ denotes element-wise addition. The switcher embedding and time embedding are injected into the diffusion model in the ResNet layers; the CLIP embedding is injected through the cross attention mechanism.}
    \label{fig:model_arch}

\end{figure*}

We detail the processing of each conditioning input and their integration into the video diffusion model, as illustrated in Figure \ref{fig:model_arch}. Starting with a single reference image \( I_{ref} \in \mathbb{R}^{H\times W\times C} \), we extract its VAE latent representation, repeat it \( T \) times, and concatenate it with the input noise latent along the channel dimension. This combined representation is passed through a convolutional layer to generate \( C_\text{vae} \in \mathbb{R}^{T\times H_1\times W_1\times C_1} \).

For the geometry cue, represented as a sequence of SMPL normal maps, we use a 2D ConvNet \( \boldsymbol{\epsilon}_{\text{geo}} \) to extract features \( C_{\text{smpl}} \in \mathbb{R}^{T\times H_1\times W_1\times C_1} \). Similarly, for the appearance cue, which consists of corresponding NeRF renderings, we pass the sequence through a VAE and subsequently a 2D ConvNet \( \boldsymbol{\epsilon}_{\text{appr}} \) to obtain features \( C_{\text{nerf}} \in \mathbb{R}^{T\times H_1\times W_1\times C_1} \).

The feature representations \( C_\text{vae} \), \( C_{\text{smpl}} \), and \( C_{\text{nerf}} \) are element-wise added and fed into the diffusion UNet \( \mathcal{U}_\Theta \) to predict noise. Additionally, the CLIP embedding of the reference image, \( \bm{h}_\text{clip} \in \mathbb{R}^d \), is repeated \( T \) times to match the frame sequence and injected into \( \mathcal{U}_\Theta \) via cross-attention.

Finally, the switcher, represented as a one-hot vector, is embedded and element-wise added to the time embedding. This is also repeated \( T \) times to align with the frame sequence and injected into \( \mathcal{U}_\Theta \) within the ResNet layers.

\subsection*{C.2 Training details}
We leverage a combination of 3D human scans, multi-view videos and monocular videos to train our diffusion UNet \( \mathcal{U}_\Theta \), geometry cue encoder \( \boldsymbol{\epsilon}_{\text{geo}} \) and appearance cue encoder \( \boldsymbol{\epsilon}_{\text{appr}} \).

For 3D scans, we utilize 20 renderings for training the novel view synthesis task. A view is randomly selected as the reference, and the model is tasked with predicting all 20 consecutive novel views. Note that the reference view and the predicted starting view do not need to be the same. Furthermore, the order of the predicted views is randomly determined, \textit{i.e.}, either clockwise or counterclockwise. 

For multi-view videos, we train the model for the novel pose synthesis task. A frame is randomly selected as the reference image, and the model is tasked with predicting a 20-frame video clip. For a single pose, we also experiment with the novel view synthesis task. However, due to the fluctuating camera trajectory and sparse camera setup, we observe suboptimal outcomes.

For monocular videos, each video is split into a sequence of images at 30 frames per second. We sample one image every four consecutive frames for training. Similarly, a frame is randomly chosen as the reference, and the model is tasked with predicting 20 consecutive frames.

\section*{D Additional results}
\subsection*{D.1 Runtime at inference}
Given a single in-the-wild image, we report the runtime of our proposed method, in terms of both novel view and pose synthesis. The detailed runtime breakdown is as follows. (1) geometry cue rendering; (2) appearance cue rendering; (3) video diffusion inference. The runtime is reported on a single NVIDIA A800 GPU and measured in seconds. 

\noindent\textbf{Novel view synthesis.} We report the runtime for generating 20 novel views given a single input image, as shown in Table \ref{tab:runtime_nvs}. 

\begin{table}[ht]
\centering
\begin{tabular}{lcccc}
\hline
 & 20 frames & Per frame \\
\hline
Geo. cue rendering & 5.81 & 0.29 \\
Appr. cue rendering & 28.6   & 1.43 \\
Diffusion inference & 15.88  & 0.79 \\ 
\hline
Total runtime & 50.29 & 2.51 \\
\hline
\end{tabular}
\caption{\textbf{Runtime at inference for generating 20 novel views of a single human image.}}
\vspace{-1.0em} 
\label{tab:runtime_nvs}
\end{table}

\noindent\textbf{Novel pose synthesis.} We report the runtime for generating 100 consecutive novel poses from a single input image, as shown in Table \ref{tab:runtime_np}. The additional video diffusion inference time arises due to the 6-frame overlap between consecutive video segment windows.

\begin{table}[ht]
\centering
\begin{tabular}{lcccc}
\hline
 & 100 frames & Per frame \\
\hline
Geo. cue rendering & 29 & 0.29 \\
Appr. cue rendering & 143 & 1.43  \\
Diffusion inference & 106.49    & 1.06  \\ 
\hline
Total runtime & 278.49 & 2.78\\
\hline
\end{tabular}
\caption{\textbf{Runtime at inference for generating 50 consucutive novel poses from a single human image.}}
\vspace{-1.0em} 
\label{tab:runtime_np}
\end{table}

\noindent\textbf{Efficiency comparison.} We provide a comparison of runtime and VRAM usage with the strongest baseline Champ \cite{zhu2024champ}, as shown in Table \ref{tab:efficiency-comparison}.

\begin{table}[h]
\centering
\begin{tabular}{lcccc}
\hline
 & fps & VRAM (GB) \\
\hline
Champ & 0.57  & 9.88 \\
Ours & 0.40 & 5.32  \\
\hline
\end{tabular}
\caption{\textbf{Efficiency comparison with baseline method.}}
\vspace{-1.0em} 
\label{tab:efficiency-comparison}
\end{table}

\subsection*{D.2 Ablation on merging novel view \& pose tasks}
We provide quantitative ablation analysis on merging both novel view and novel pose tasks into one model. From the static novel view synthesis perspective, the unified framework allow us to train on the abundant internet videos - which leads to the improved generalization, as shown in the main paper. It does not hurt the in-domain dataset performance according to Table \ref{tab:merge_thuman_2k2k}.

From the pose animation perspective, training on additional 3D datasets can improve the quality when we animate the avatar from a novel view, as shown in Table \ref{tab:ablation_quantitative_merging}.

\begin{table}[h]
\centering
\resizebox{0.8\columnwidth}{!}{%
\begin{tabular}{lcccc}
\hline
Training data & PSNR $\uparrow$ & SSIM $\uparrow$ & LPIPS $\downarrow$ & FVD $\downarrow$ \\
\hline
Monocular videos & 28.67  & \textbf{0.946} & 0.041  & 208.3 \\
Full & \textbf{28.74} & 0.945 & \textbf{0.040}  & \textbf{188.5} \\

\hline
\end{tabular}
}
\caption{\textbf{Quantitative ablation on merging view synthesis and pose animation into one model.} Results reported for novel view animation task on MVHumanNet dataset.}
\label{tab:ablation_quantitative_merging}

\end{table}

\subsection*{D.3 Ablation on switcher}
In addition to the qualitative ablation for the switcher presented in the main paper, we also conduct quantitative ablation analysis on the switcher. With the switcher, our method effectively supports both novel view and pose synthesis, while also providing comparable or even better quantitative results, as shown in Table \ref{tab:ablation_switcher}.

\begin{table}[h]
\centering
\resizebox{0.8\columnwidth}{!}{%
\begin{tabular}{lcccc}
\hline
 & PSNR $\uparrow$ & SSIM $\uparrow$ & LPIPS $\downarrow$ & FVD $\downarrow$ \\
\hline
w.o. switcher & 26.58 & \textbf{0.944} & 0.042 & 198.9 \\
w. switcher & \textbf{26.77} & 0.943 & \textbf{0.041}  & \textbf{194.8} \\

\hline
\end{tabular}
}
\caption{\textbf{Ablation on switcher for novel view synthesis task on THuman.}}
\label{tab:ablation_switcher}
\end{table}

\subsection*{D.4 Novel pose results on scan datasets}
\begin{figure}[h]
    \centering
    \includegraphics[width=0.45\textwidth]{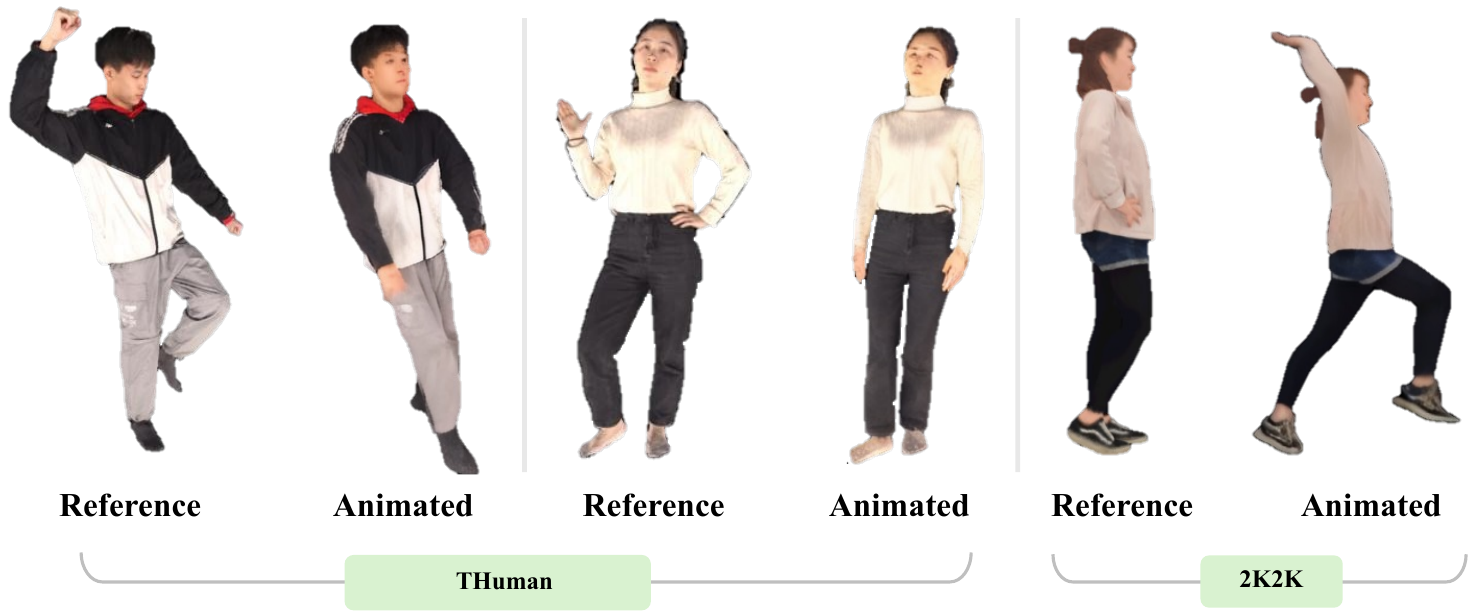}
    \caption{\textbf{Novel pose results on THuman and 2K2K.} The reference images are animated by pose sequences derived from MVHumanNet dataset.}
    \label{fig:thuman_2k2k_animation}
\end{figure}

We show the novel pose animation results for subjects in THuman and 2K2K in Figure \ref{fig:thuman_2k2k_animation}, where the reference images are animated through pose sequences derived from disparate videos.

\begin{table*}[ht]
\centering
\resizebox{0.7\textwidth}{!}{%
\begin{tabular}{lcccccccc}
\toprule
Training data & \multicolumn{2}{c}{PSNR $\uparrow$} & \multicolumn{2}{c}{SSIM $\uparrow$} & \multicolumn{2}{c}{LPIPS $\downarrow$} & \multicolumn{2}{c}{FVD $\downarrow$} \\
\cmidrule(lr){2-3} \cmidrule(lr){4-5} \cmidrule(lr){6-7} \cmidrule(lr){8-9}
& THuman & 2K2K & THuman & 2K2K & THuman & 2K2K & THuman & 2K2K \\
\midrule
3D scans only & \textbf{26.82}  & 28.76 & 0.936  & 0.953  & \textbf{0.040}  & 0.040  & \textbf{189.5}   & \textbf{187.9} \\
3D scans+dynamic videos & 26.77 & \textbf{28.82}  & \textbf{0.943}  & \textbf{0.954} & 0.041 & \textbf{0.039}   & 194.8 & 191.3 \\
\bottomrule
\end{tabular}
}
\caption{\textbf{Quantitative ablation on merging view synthesis and pose animation into one model.} Results reported for novel view synthesis task on in-domain THuman and 2K2K testset. 
}
\label{tab:merge_thuman_2k2k}
\end{table*}

\subsection*{D.5 Robustness to input view angles}
\begin{figure*}[t]
    \centering
\includegraphics[width=0.9\textwidth]{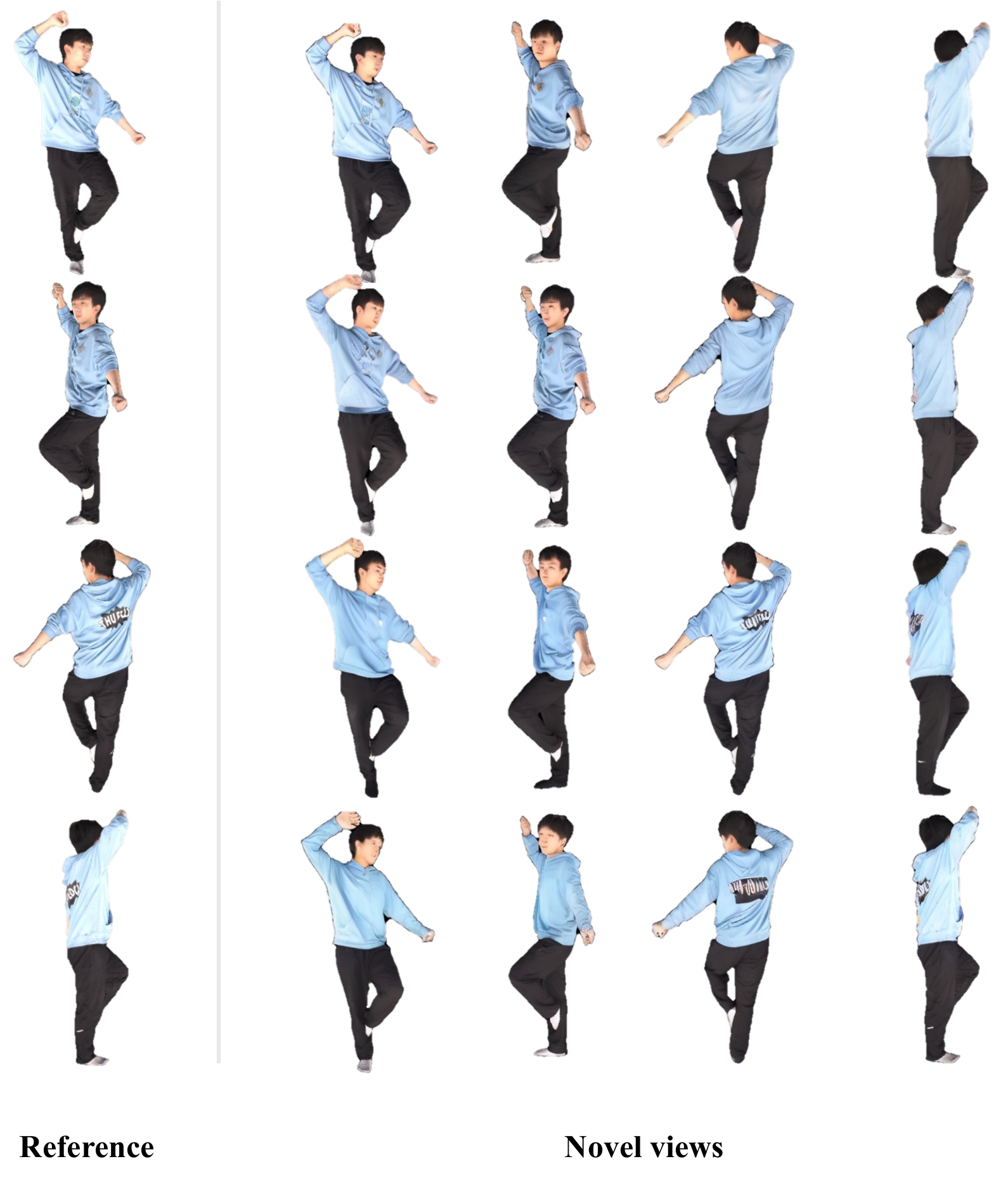}
    \caption{\textbf{Qualitative results of the generated novel views for various input views of the same human subject.} }
    \label{fig:view_robustness}
\end{figure*}

We train our model using an arbitrary view as input.
Thus, we are interested in the novel view synthesis robustness from different input view angles. To this end, we present our quantitative results in Table \ref{tab:nvs_robustness}. We find that our pipeline demonstrates robustness with different views of input. Visualized results are shown in Figure \ref{fig:view_robustness}, where our proposed method can maintain
the faithful appearance near the reference view and generate reasonable appearances in unseen regions.

\begin{table}[h]
\centering
\resizebox{0.9\columnwidth}{!}{%
\begin{tabular}{lcccc}
\hline
Angle & PSNR $\uparrow$ & SSIM $\uparrow$ & LPIPS $\downarrow$ & FVD $\downarrow$ \\
\hline
Front & 28.46   & 0.952  & 0.04  & 178.3\\
Back & 28.34   & 0.952   & 0.04  & 211.5 \\
Left side & 29.18  & 0.956  & 0.039  & 199.9  \\
Right side & 29.29   & 0.955   & 0.038  & 175.2  \\
\hline
Mean & 28.82 & 0.954 & 0.039 & 191.3 \\
Std & 0.487 & 0.002 & 0.001 & 17.42 \\
\hline
\end{tabular}
}
\caption{\textbf{Quantitative results on 2K2K dataset for novel view synthesis from different input view angles.}}
\vspace{-1.0em} 
\label{tab:nvs_robustness}
\end{table}

\subsection*{D.6 Ablation on video diffusion model }
\begin{figure}[h]
\centering

\begin{minipage}{0.49\linewidth}
\centering
\resizebox{\linewidth}{!}{%
\begin{tabular}{lcccccccc}
\toprule
 & \multicolumn{2}{c}{PSNR $\uparrow$} & \multicolumn{2}{c}{SSIM $\uparrow$} & \multicolumn{2}{c}{LPIPS $\downarrow$} & \multicolumn{2}{c}{FVD $\downarrow$} \\
\cmidrule(lr){2-3} \cmidrule(lr){4-5} \cmidrule(lr){6-7} \cmidrule(lr){8-9}
& NVS & NPS & NVS & NPS & NVS & NPS & NVS & NPS \\
\midrule
Before diff. & 24.25 & 18.37   & 0.925  & 0.809   & 0.073  & 0.233  & 517.7  & 1255.8 \\
After diff. & \textbf{28.82} & \textbf{19.11}  & \textbf{0.954}  & \textbf{0.833}  & \textbf{0.039}  & \textbf{0.176}   &  \textbf{191.3}   & \textbf{362.0} \\
\bottomrule
\end{tabular}
}
\end{minipage}
\hfill
\begin{minipage}{0.49\linewidth}
\centering
\includegraphics[width=\linewidth]{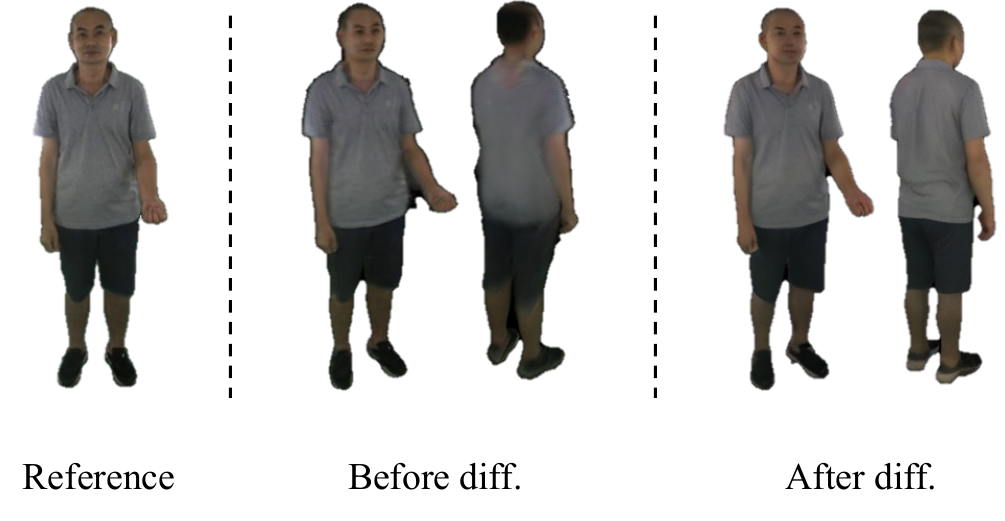}
\vspace{-1.5em}
\end{minipage}
\vspace{-0.1in}
\caption{Quantitative and qualitative results before/after diff. }
\label{fig:diff_ablation_combined}
\end{figure}

Fig.~\ref{fig:diff_ablation_combined} (left) presents quantitative ablations on both novel view and pose synthesis tasks, showing consistent improvements over the direct output of the generalizable NeRF results, denoted as before diff. Fig.~\ref{fig:diff_ablation_combined} (right) shows qualitative results on MVHumanNet dataset: NeRF helps the diffusion model preserve identity on nearby views, while the diffusion model refines distant views, producing sharper results with fewer artifacts.

\subsection*{D.7 Video results}
We provide additional video results, including in-the-wild avatar synthesis, comparison with baselines, ablations and additional results. Details explained below:

\begin{itemize}
    \item In-the-wild avatar synthesis: We demonstrate novel view synthesis, pose animation and interactive 4D video synthesis on in-the-wild avatars. 
    \item Comparison with baselines: We compare our results with Animate Anyone \cite{hu2023animateanyone} and Champ \cite{zhu2024champ} on THuman \cite{tao2021function4d} for novel view synthesis and TikTok \cite{jafarian2021learning} for pose animation. The video results demonstrate our method outperforms the two baseline methods in terms of consistency and quality.
    \item Ablations: We highlight the importance of the dense appearance cue and switcher through the ablation videos.
    \item Additional results: We show novel view synthesis results on MVHumanNet \cite{xiong2024mvhumannet}. We also show free-view animation results, where the human subject and the camera are both moving.
\end{itemize}

\section*{E Verification of design choices}
We found that generative human novel view synthesis remains relatively under-explored, primarily due to the challenge of synthesizing consistent appearances across multiple novel views, especially in unseen regions. We have shown the importance of generalizable appearance cue and geometry cue in the main paper. Here, we aim to additionally validate the choice of leveraging video diffusion models and training with smooth camera orbits.

\subsection*{E.1 Image v.s. video diffusion model}
\begin{figure}[h]
    \centering
\includegraphics[width=0.45\textwidth]{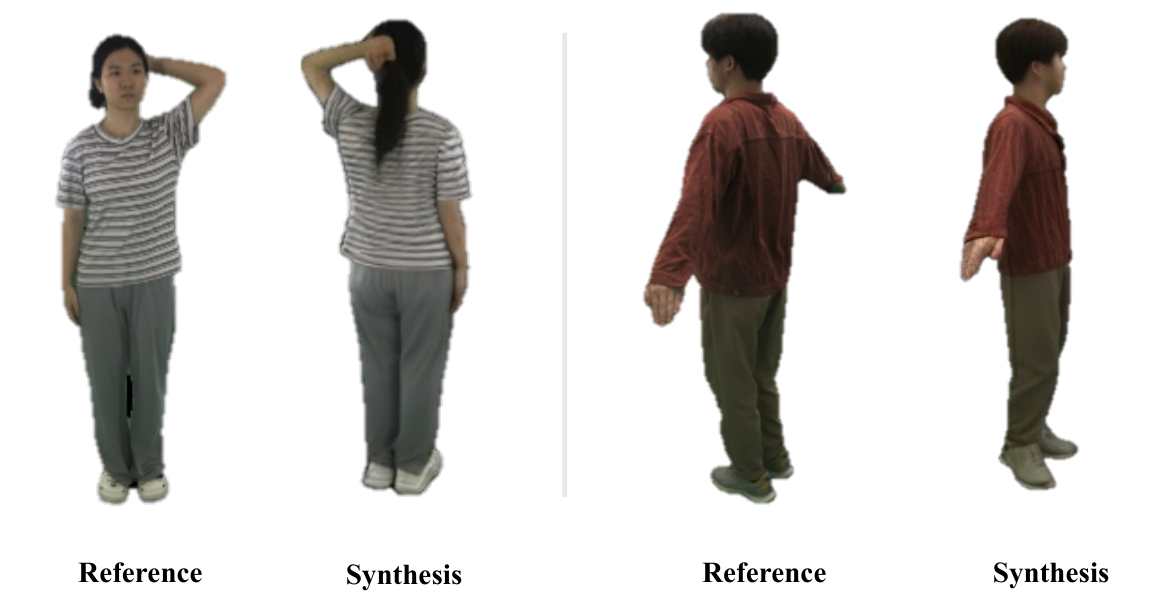}
    \caption{\textbf{Image-to-image novel view synthesis on MVHumanNet dataset by using image diffusion model and appearance cue.}}
    \label{fig:img-2-img}
\end{figure}

Image diffusion model is believed to be good at image-to-image translation tasks. Specifically, given a single reference image and the target appearance cue, we can train a image diffusion model to synthesize the target image. The model architecture can be a variant of Animate Anyone \cite{hu2023animateanyone} and Champ \cite{zhu2024champ}, where we adopt two ReferenceNets to inject the rich information from the reference image and appearance cue respectively. The results are shown in Figure \ref{fig:img-2-img}. To enable multi-view consistent synthesis for a single subject, we have tried adding an 1D temporal-axis attention layers \cite{guo2023animatediff} and only fine-tune these new added layers. We have also attempted to leverage human geometric prior to construct 3D correspondence across different views \cite{huang2024epidiff}. 

\begin{figure}[h]
    \centering
\includegraphics[width=0.45\textwidth]{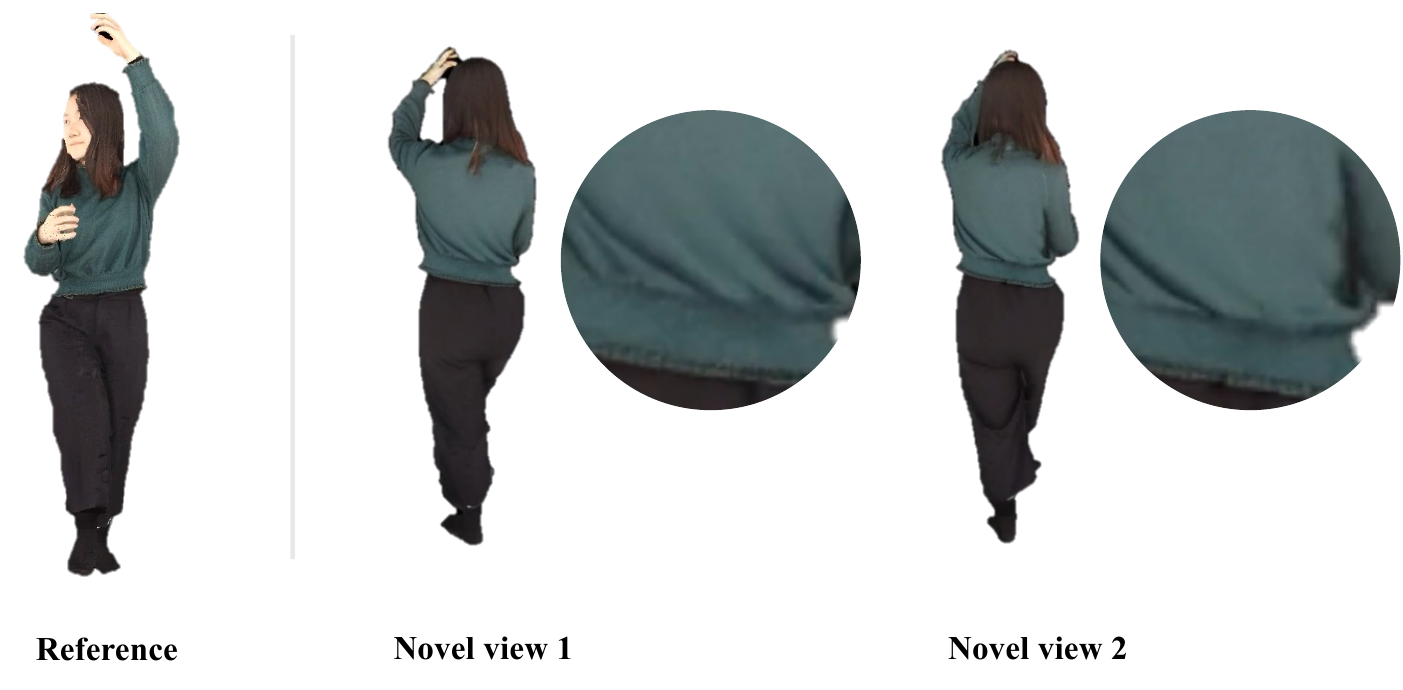}
    \caption{\textbf{Novel view synthesis on THuman2.1 by a multi-view image-based diffusion model.} Inconsistent clothing wrinkles appear between two adjacent novel view generations.}
    \label{fig:mv_diffusion_model}
\end{figure}

\noindent\textbf{Consistency comparisons.} Compared to directly using a pretrained video diffusion model, these image-based diffusion methods exhibit significantly inferior performance. Their results are shown in Figure \ref{fig:mv_diffusion_model}.

\noindent\textbf{Free-view interpolation.} Due to GPU memory limitations, we are restricted to training with approximately 20 novel views per subject per batch. During inference, we also test and compare the ability of both models to generate a dense trajectory of novel views (\textit{e.g.,} 100 views). However, neither approach achieves satisfactory results: image-based diffusion models show significant inconsistencies, while video diffusion models produce blurry frames.

\subsection*{E.2 Novel view camera trajectories}
On our 3D scan dataset, we render a smooth camera trajectory with the same elevation and evenly distributed azimuth. We also explore the possibility of leveraging the fluctuant novel views in MVHumanNet dataset to learn the multi-view consistency. However, we empirically find that the diffusion models are not able to capture the consistency even with the aid of appearance cues, causing the textures to flow across views.

\end{document}